\documentclass[letterpaper, 10 pt, conference]{ieeeconf}  % Comment this line out if you need a4paper

\IEEEoverridecommandlockouts                              % This command is only needed if 
\usepackage{cite}
\usepackage{amsmath,amssymb,amsfonts}
\usepackage{algorithmic}
\usepackage{graphicx}
\usepackage{textcomp}
\usepackage{xcolor}
\usepackage{enumerate}
\usepackage[linesnumbered,ruled,lined]{algorithm2e}
\usepackage{booktabs}
\usepackage{threeparttable}
\usepackage{multirow}
\usepackage{stfloats}
\usepackage{subfigure}
\usepackage{graphicx}
\usepackage{rotating}
\usepackage{url}
\usepackage{colortbl}  %彩色表格需要加载的宏包
\usepackage{xcolor}
\usepackage{array}   %对表列和表格线的设置需要用到array宏包
\usepackage{geometry}
\usepackage{hyperref}
\title{\LARGE \bf
BSH-Det3D: Improving 3D Object Detection with BEV Shape Heatmap
}

\author{You Shen$^{1}$, Yunzhou Zhang$^{1*}$, Yanmin Wu$^{2}$, Zhenyu Wang$^{3}$, Linghao Yang$^{1}$, \\Sonya Coleman$^{4}$, Dermot Kerr$^{4}$
	\thanks{$^*$The corresponding author of this paper. }
	\thanks{$^{1}$You Shen, Yunzhou Zhang, Linghao Yang are with College of Information Science and Engineering, Northeastern University, Shenyang 110819, China (Email: {\tt\small zhangyunzhou@mail.neu.edu.cn}).}
         \thanks{$^{2}$Yanmin Wu is with School of Electronic and Computer Engineering, Peking University, Shenzhen, China.}
         \thanks{$^{3}$Zhenyu Wang is with Faculty of Robotics and Engineering of Northeastern University, Shenyang, China, and the Department of Electronic and Computer Engineering of Technical University in Munich, Germany.}
         \thanks{$^{4}$Sonya Coleman and Dermot Kerr are with School of Computing, Engineering and Intelligent Systems, Ulster University, N. Ireland, UK.}
	\thanks{This work was supported by National Natural Science Foundation of China (No. 61973066), Major Science and Technology Projects of Liaoning Province(No. 2021JH1/10400049),   Fundamental Research Funds for the Central Universities(N2004022).}
}

\begin{document}

\maketitle
\thispagestyle{empty}
\pagestyle{empty}

%%%%%%%%%%%%%%%%%%%%%%%%%%%%%%%%%%%%%%%%%%%%%%%%%%%%%%%%%%%%%%%%%%%%%%%%%%%%%%%%
\begin{abstract}
The progress of LiDAR-based 3D object detection has significantly enhanced developments in autonomous driving and robotics. However, due to the limitations of LiDAR sensors, object shapes suffer from deterioration in occluded and distant areas, which creates a fundamental challenge to 3D perception. Existing methods estimate specific 3D shapes and achieve remarkable performance. However, these methods rely on extensive computation and memory, causing imbalances between accuracy and real-time performance. To tackle this challenge, we propose a novel LiDAR-based 3D object detection model named BSH-Det3D, which applies an effective way to enhance spatial features by estimating complete shapes from a bird's eye view (BEV). Specifically, we design the Pillar-based Shape Completion (PSC) module to predict the probability of occupancy whether a pillar contains object shapes. The PSC module generates a BEV shape heatmap for each scene. After integrating with heatmaps, BSH-Det3D can provide additional information in shape deterioration areas and generate high-quality 3D proposals. We also design an attention-based densification fusion module (ADF) to adaptively associate the sparse features with heatmaps and raw points. The ADF module integrates the advantages of points and shapes knowledge with negligible overheads. Extensive experiments on the KITTI benchmark achieve state-of-the-art (SOTA) performance in terms of accuracy and speed, demonstrating the efficiency and flexibility of BSH-Det3D. The source code is available on \url{https://github.com/mystorm16/BSH-Det3D}.
\end{abstract}

%%%%%%%%%%%%%%%%%%%%%%%%%%%%%%%%%%%%%%%%%%%%%%%%%%%%%%%%%%%%%%%%%%%%%%%%%%%%%%%%
\section{INTRODUCTION}

Over the past decade, deep learning has made significant progress in 2D vision tasks such as detection \cite{girshick2015fast,liu2016ssd,redmon2016you}, segmentation\cite{long2015fully,he2017mask,liu2018path}, and pose estimation\cite{chen2018cascaded}. While 2D images have valuable information, 3D point clouds can provide more geometric, shape and scale information\cite{guo2020deep}, significantly improving scene perception capability. LiDAR sensors have been widely used to obtain 3D point clouds in autonomous driving, mobile robotics, and augmented reality/virtual reality thanks to high-precision measurements and robustness to illumination changes. Although LiDAR sensors have these advantages, achieving high-performance detection in point clouds is still challenging due to two inherent limitations:

\begin{itemize}
	\item Laser beams return after hitting the first object, causing shapes behind the occluder to be missing.
	\item The faraway objects receive only a few points on their surfaces, so shapes at far-range areas will be sparse and incomplete.
	%The source code will be released to the community.
\end{itemize}
\begin{figure}[!t]
    \vspace{2mm}
	\centering
	\includegraphics[scale=0.47]{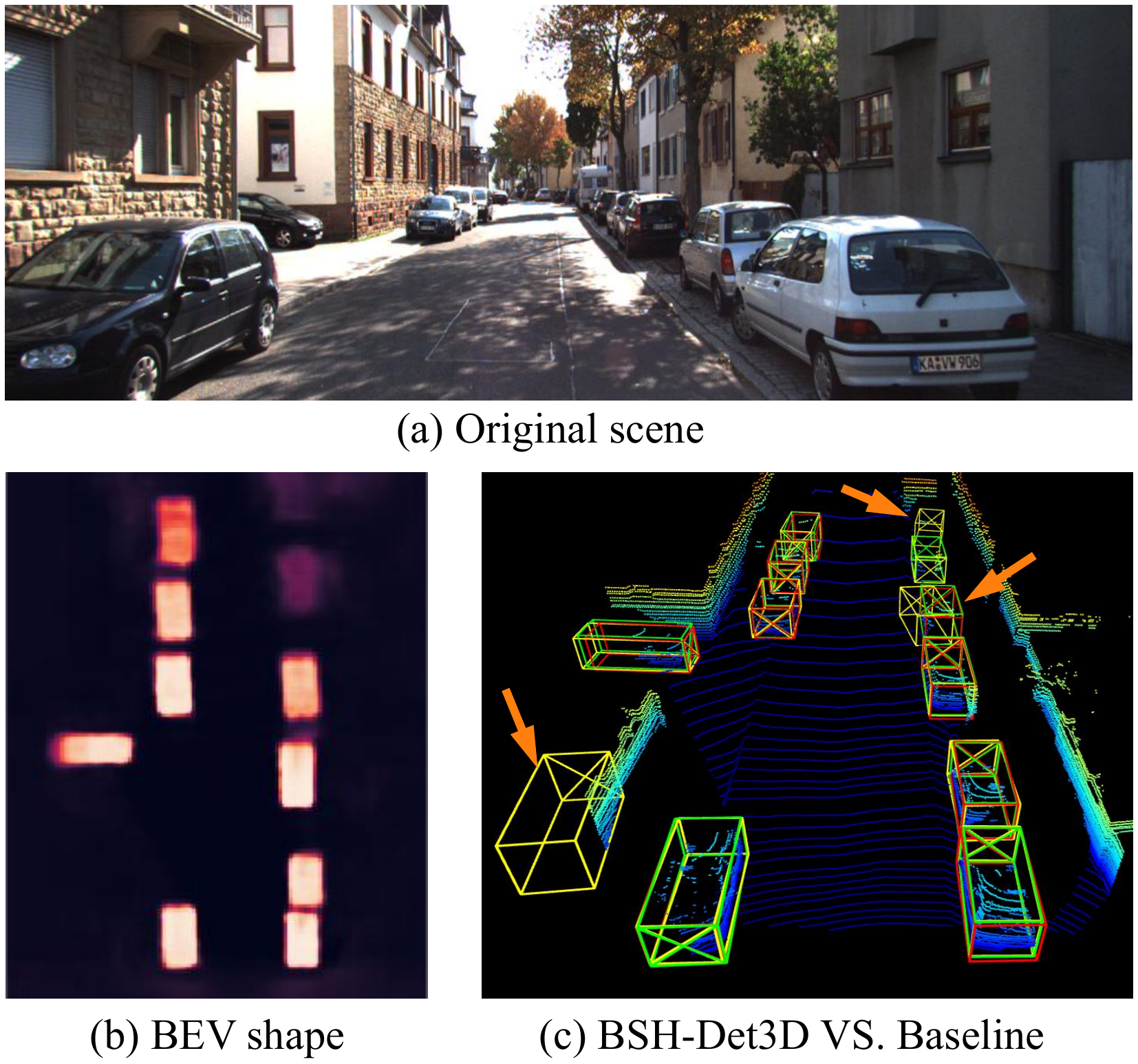}
	\vspace{-3mm}
	\caption{Performance comparisons of our BSH-Det3D with the baseline\cite{lang2019pointpillars} on the KITTI \textit{val} set. (a) RGB image of the original scene. (b) Result of BEV shape heatmap. (c) Result of our BSH-Det3D and the baseline detector. The ground-truth boxes, predicted boxes of the baseline and BSH-Det3D are shown in red, yellow, and green. As indicated by the orange arrows, our method can fix box offset and removes false positives effectively.}
	\label{P1}
	\vspace{-4mm}
\end{figure}

Both limitations cause shape deterioration during detection. To address this issue, Xu \textit{et al}.\cite{xu2022behind} complete entire shapes by manual ground truth. Subsequently, almost all objects are detected correctly (Average Precision\textgreater99$\%$), proving that complete shapes are essential for high-performance detectors. One straightforward way to alleviate the missing shape problem is learning object shapes from priors: SPG\cite{xu2021spg} and BtcDet\cite{xu2022behind} manually fill points into the labeled bounding boxes and try to recover the specific shape of objects. However, since they focus on entire 3D shapes, these methods are computationally expensive. Other methods convert shape information to features instead of specific shapes\cite{du2020associate,shi2020points,he2020structure} to reduce computation. However, this is challenging due to shape feature extraction and fusion resulting lower accuracy. Thus, a new challenge arises: \textbf{How do detectors alleviate shape deterioration while remaining efficient and flexible? }

\begin{figure}[ht]
    \vspace{0mm}
	\centering
	\includegraphics[scale=0.18]{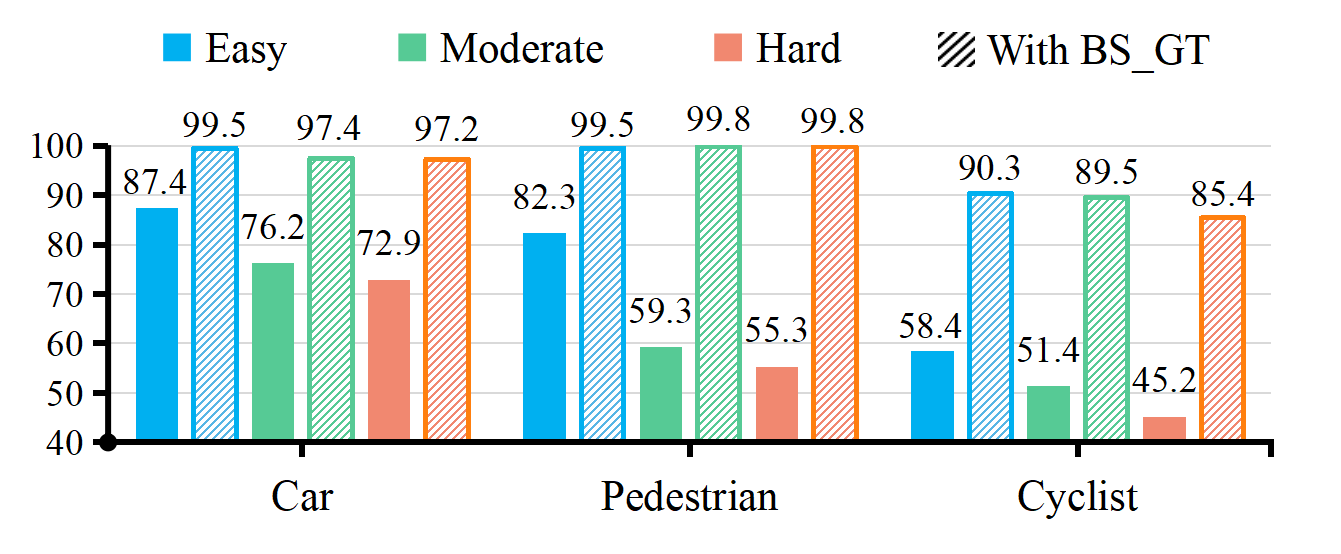}
	\vspace{-6mm}
	\caption{This shows the performance of vanilla PointPillars\cite{lang2019pointpillars} 
 (solid bar) and PointPillars with ground truth BEV shape heatmap (striped bar) on the KITTI\cite{geiger2012we} \textit{val} split. We show the result in three classes: Car, Pedestrian, and Cyclist; with three difficulty levels: Easy, Moderate, and Hard. After associating with the BEV shape, the performance significantly improves. }
	\label{P2}
	\vspace{-7.1mm}
\end{figure}

To tackle this challenge, we chose the BEV-based method for its speed and accuracy performance, \textit{e.g.}, PointPillars\cite{lang2019pointpillars}. These methods encode points into pseudo-images and use well-established 2D convolutions to extract features. We conduct a pilot study to explore the possibility of detectors with BEV shapes. We design ground truth BEV shapes as heatmaps (see details in \ref{section:A}). Then, we directly concatenate heatmaps with raw point features extracted by the baseline detector\cite{lang2019pointpillars}. Both training and evaluation is based on the KITTI\cite{geiger2012we} dataset. We show the average precision (AP) of cars, pedestrians, and cyclists with three occlusion levels. As illustrated in Fig.\ref{P2}, the performance improves significantly with the fusion of BEV shape heatmaps, demonstrating the potential of utilizing BEV shapes to enhance detection.

In this paper, we present a novel approach to improving 3D object detectors with BEV Shape Heatmap (BSH-Det3D), which alleviates shape deterioration efficiently, as illustrated in Fig.\ref{P1}. BSH-Det3D proposes an effective method for improving detection quality by learning BEV shape knowledge. We design a pillar-based shape completion module (PSC) to obtain a BEV shape heatmap in each scene. PSC extracts multi-scale pillar features and estimates the occupancy probability of whether a pillar belongs to complete shapes. Furthermore, targeting the sparsity of both point cloud features and shape heatmaps, we propose an attention-based densification fusion module (ADF) to associate points and shapes. We conduct experiments on the KITTI dataset, and achieve SOTA performance in terms of accuracy and speed. It is worth noting that our method is general and can be used with various detectors. To validate the flexibility, we conduct experiments integrating our method into different mainstream 3D detection frameworks\cite{lang2019pointpillars,yan2018second} and design a two-stage module for box refinement.

The main contributions can be summarized as follows:
\begin{itemize}
	\item We propose a novel 3D object detector that learns to associate object shape knowledge from a bird's eye view, enhancing spatial features and providing implicit guidance for detection.
	\item We design a pillar-based shape completion module (PSC) to estimate a probability-based shape heatmap for each scene, alleviating shape deterioration efficiently.
        \item We design an attention-based densification fusion module (ADF) that adapts to the sparse features of shapes and point clouds with negligible overheads.
        \item Our proposed BSH-Det3D achieves SOTA performances on the KITTI benchmark in terms of accuracy and real-time performance. We also test our approach with different baseline detectors to verify its flexibility.
	%The source code will be released to the community.
\end{itemize}

 \section{RELATED WORK}
 \subsection{LiDAR-based 3D Object Detectors}
 According to the representation of point clouds, LiDAR-based 3D object detection can be divided into point-based and grid-based methods. Point-based methods inherit the success of feature extraction modules\cite{qi2017pointnet} and propose diverse architectures to detect objects from raw points directly. PointRCNN\cite{shi2019pointrcnn} segments point clouds by PointNet++\cite{qi2017pointnet++} and estimates proposals for each foreground point. STD\cite{yang2019std} presents a point-based proposal generation paradigm with spherical anchors to reduce computation. 3DSSD\cite{yang20203dssd} combines feature-based and point-based sampling to improve the classification.
Grid-based 3D detectors first transfer raw points into discrete grid representations such as voxels and pillars; then, detectors use 2D or 3D convolutional neural networks to extract features from grids and detect objects from grid cells. VoxelNet\cite{zhou2018voxelnet} divides point cloud into 3D voxels, which are further processed by the voxel feature extractor and 3D CNN encoder network. SECOND\cite{yan2018second} introduces sparse 3D convolutions for efficient 3D processing of voxels, significantly improving real-time performance. PointPillars\cite{lang2019pointpillars} collapses raw points into vertical pillars, uses a per-pillar feature extractor by PointNet\cite{qi2017pointnet} to compress the height dimension, and then utilizes the pillar feature as a BEV pseudo-image for detection. Unlike existing BEV detectors which encode points into pseudo-image and directly estimate proposals, we propose a novel way of predicting shape heatmaps in BEV to enhance spatial features. Our BSH-Det3D is designed on grid-based methods and can be easily integrated into various detectors.
 \subsection{Shape Priors for 3D Object Detection}
 Many advanced 3D detectors focus on alleviating missing shape problems by using shape priors. One class of methods attempts to learn shape knowledge as features: Part-A²\cite{shi2020points} applies a part-aware stage to obtain object part locations. SA-SSD\cite{he2020structure} and Associate-3Ddet\cite{du2020associate} aim to exploit structure modules to conserve shape features by auxiliary networks. However, due to difficulties in extracting and fusing shape features, these methods suffer from accuracy bottlenecks. Another class of methods recovers 
 the specific shape of objects: SPG\cite{xu2021spg} locates foreground regions and generates semantic points for each foreground voxel. BtcDet\cite{xu2022behind} finds similar shapes as priors and generates new shape points. Due to the estimate of entire 3D shapes, these methods consume significant computation and memory, resulting in imbalances between accuracy and real-time performance. Given this, BSH-Det3D proposes a novel mechanism for utilizing shape priors as BEV heatmaps, which effectively reduces shape deterioration with a small amount of extra time, particularly for occluded objects or distant objects.

\section{METHODOLOGY}
\subsection{Model Overview}
\begin{figure*}[ht]
	\centering
	\includegraphics[scale=0.78]{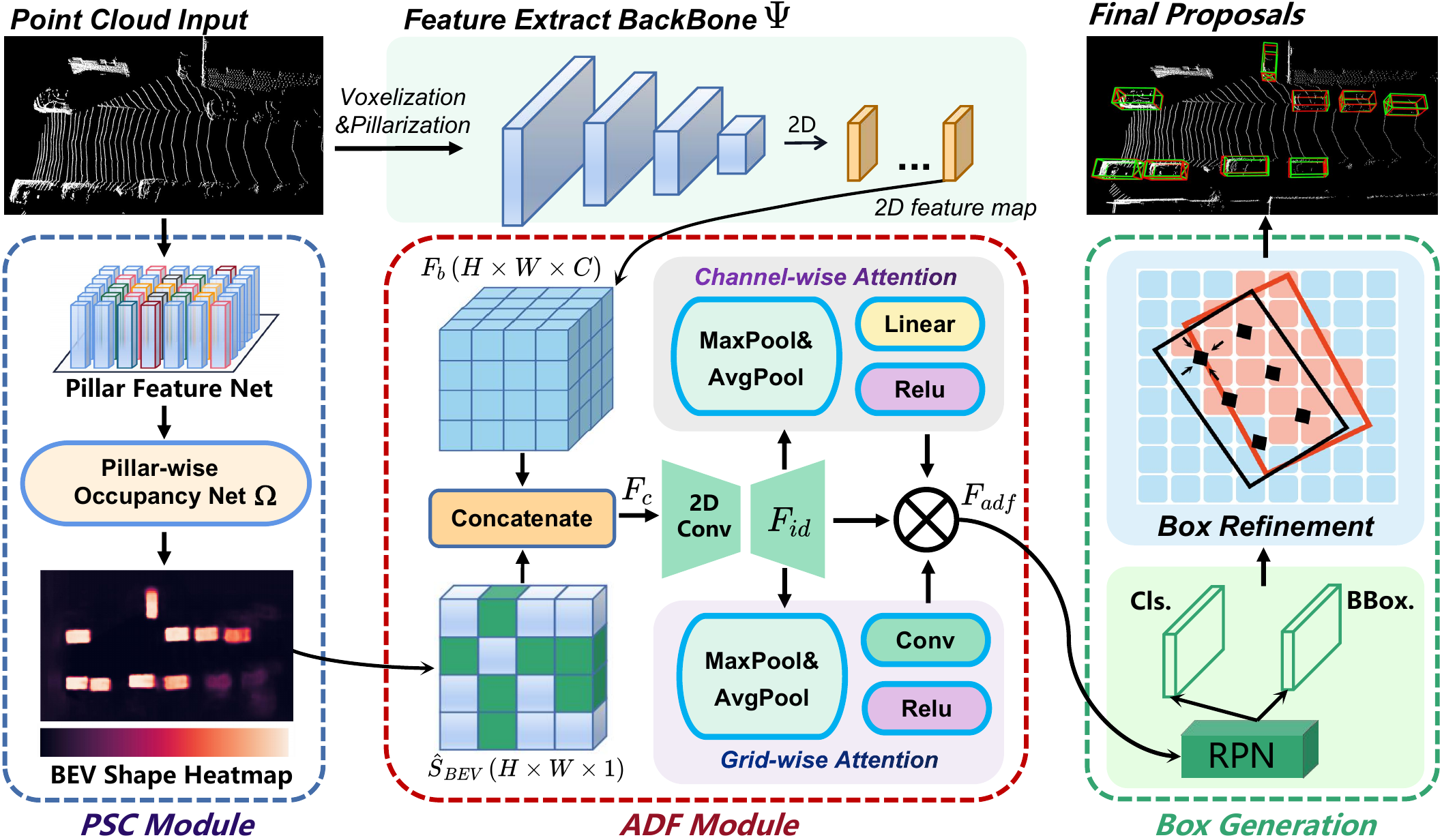}
	\vspace{-3mm}
	\caption{\textbf{Detection pipeline:} The PSC module first splits points into pillars and estimates pillar-wise shape occupancy probability to generate BEV shape heatmap $\hat{S} _{BEV}$. The backbone network $\Psi$ extracts feature from raw points, and $\hat{S} _{BEV}$ concatenates with the output feature of $\Psi$. Then, the ADF module uses a hybrid-attention based strategy to fuse features, linking with an RPN network to generate 3D proposals. Furthermore, for
  each proposal, BSH-Det3D constructs local grids and pools the local features with $\hat{S} _{BEV}$ to the nearby grids for further refinement (see black box and red box in Box Refinement).}
	\label{P3}
	\vspace{-6mm}
\end{figure*}

As illustrated in Fig.\ref{P3}, the PSC module first pillarizes raw points and utilizes a pillar-wise occupancy network to estimate the BEV shape heatmap $\hat{S} _{BEV}$ (\ref{section:A}). 
Next, BSH-Det3D uses a backbone network $\Psi$\cite{yan2018second}\cite{lang2019pointpillars} to extract features ${F}_{b}$ of point clouds. To associate the shape knowledge, the ADF module concatenates the $\hat{S} _{BEV}$ to the output feature from $\Psi$ and gets $F_{c}$; we then use a hybrid-attention based strategy to get fusion features ${F} _{adf}$.
Finally, ${F} _{adf}$ is sent to a Region Proposal Network (RPN) generating 3D proposals. During box refinement, we construct local grids covering each proposal box, and aggregate the grid features with shape heatmaps to generate the final proposals.

\subsection{Pillar-based Shape Completion Module}\label{section:A}
\begin{figure}[ht]
    \vspace{0mm}
	\centering
	\includegraphics[scale=0.52]{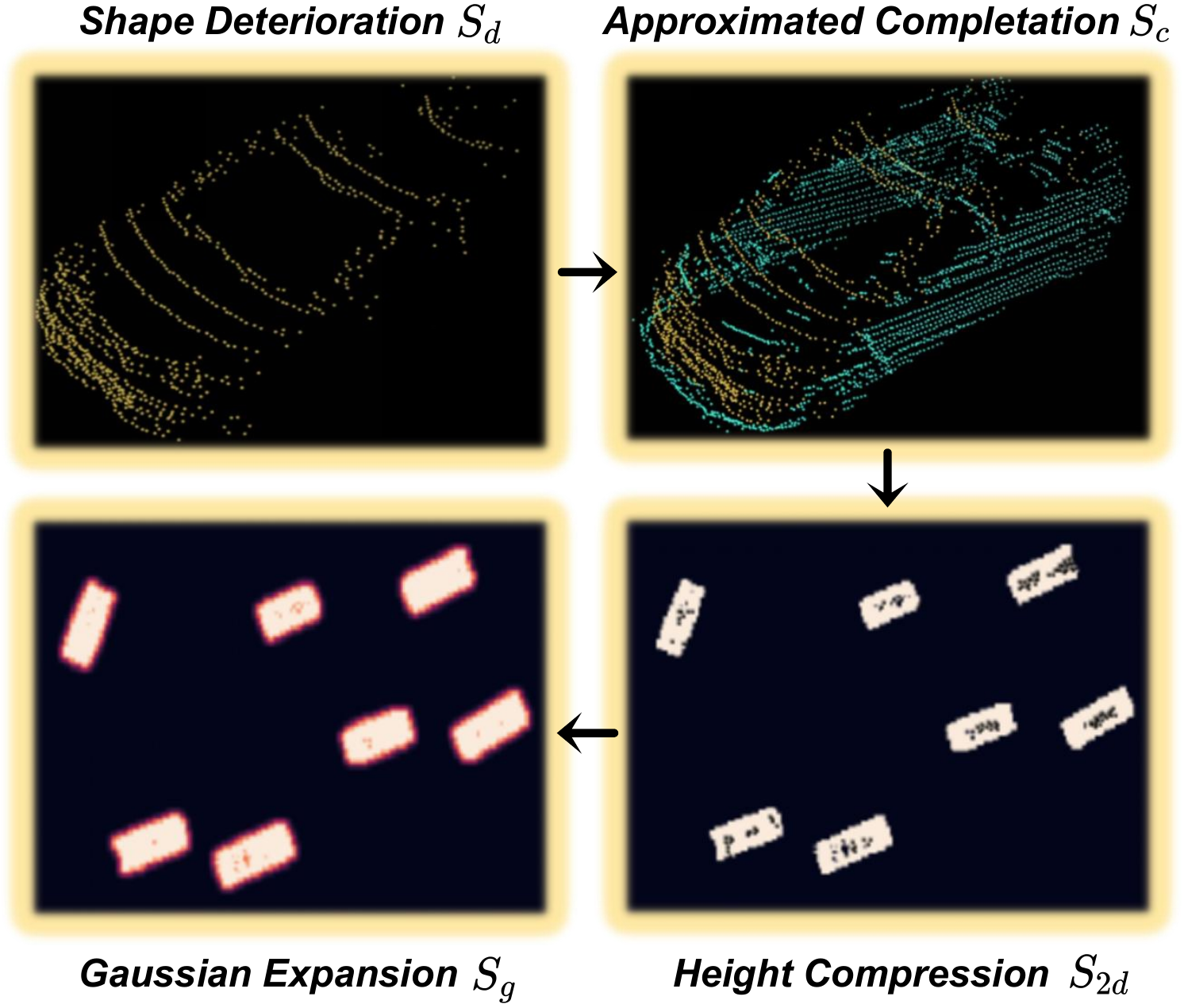}
	\vspace{-3mm}
	\caption{The generation process of our BEV shape label. For each shape deteriorated object $S_{d}$, we obtain completed 3D shapes by following the method in\cite{xu2022behind}. Next, we compress $S_{c}$ into 2D by height dimension, getting $S_{2d}$. After that, we apply rendered Gaussian kernels to counteract the noise introduced by the shape priors, getting high-quality BEV shape labels $S_{g}$.}
	\label{P4}
	\vspace{-6mm}
\end{figure}

\textbf{Generation of ground truth labels.} The process of generating BEV shape labels is illustrated in Fig.\ref{P4}. First, we follow methods in\cite{xu2022behind} using a heuristic-based strategy to find the top three shapes which are most similar to current shapes $S_{d}$ in the dataset. After assembling similar shapes, the approximate 3D shapes $S_{c}$ can be obtained. Next, we compress the height dimension of $S_{c}$ and get the corresponding 2D shapes $S_{2d}$. In $S_{2d}$, we set shape occupancy probability $P(O_{s} )=1$ for pillars that contain shapes and $P(O_{s} )=0$ for the others. When multiple objects are assembled as $S_{c}$, the point density varies, which causes voids in $S_{2d}$. To counteract this, we increase the positive supervision for the $S_{2d}$ by span locations $p_{xy}$ where $P(O_{s} )=1$ with a Gaussian kernel:
\vspace{0mm}
\begin{equation}
\begin{split}
Y_{xy} =exp(-\frac{(x-p_{x} )^{2}+(y-p_{y})^{2}}{2\sigma_{p} ^{2}} ),
\end{split}\vspace{0mm}
\end{equation}where $\sigma_{p}$ is an object size-adaptive standard deviation depending on the size of the object. Finally, $S_{g}$ is used as the ground truth label of the BEV shape heatmap.
% k\left(\mathbf{p}, \mathbf{p}^{\prime}\right)=\mathrm{e}^{-\frac{\left\|\mathbf{p}-\mathbf{p}^{\prime}\right\|^2}{2 \sigma_p^2}} 

\begin{figure}[ht]
    \vspace{0mm}
	\centering
	\includegraphics[scale=0.46]{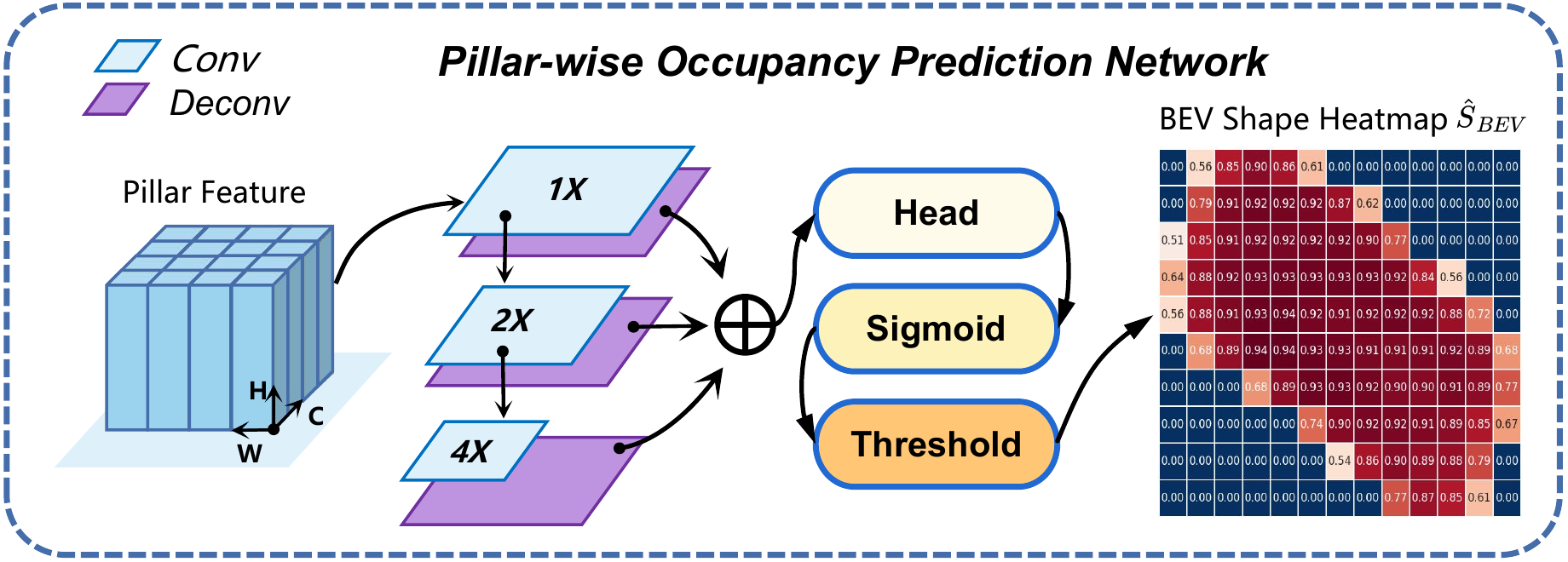}
	\vspace{-6.3mm}
	\caption{The estimation of the BEV shape heatmap. We adopt a top-down architecture to extract multi-scale pillar features. The detection head predicts the shape occupancy probability as a heatmap. After sending it to the sigmoid function and threshold, we can get the finaly heatmap $\hat{S} _{BEV}$.}
	\label{P5}
        \vspace{-6mm}
\end{figure}

\textbf{Estimation of BEV shape heatmap.} First, we adopt the strategy of \cite{lang2019pointpillars} to extract pillar features: PSC module projects raw points on the X-Y plane via a tiny one-layer PointNet\cite{qi2017pointnet} to fetch pillar feature $f_{p}$. Then, $f_{p}$ is processed by a pillar-wise shape occupancy network $\Omega $ (Fig.\ref{P5}). $\Omega $ adopts a top-down architecture that processes the $f_{p}$ with stride $1\times$, $2\times$, and $4\times$ convolution blocks, each block linking to a transposed 2D convolution for upsampling and then concatenating the multi-scale features for the detection head. The detection head uses $3\times 3 $ convolutional layers separated by ReLU and BatchNorm. The last convolutional layer produces a K-channel heatmap $\hat{Y}$ showing the shape occupancy probability, where the channel of $\hat{Y}$ indicates object classes. To highlight the shapes, we use a sigmoid function with the threshold$(\ge0.5)$ to filter $\hat{Y}$. Finally, we get the estimated BEV shape heatmap $\hat{S} _{BEV}$.

Sigmoid cross-entropy Focal Loss\cite{lin2017focal} supervises the output of $\Omega $, if $Y_{x y }=1$: 
\vspace{-2mm}
\begin{equation}\label{eq2}
    L_{shape}=-\frac{1}{N} \sum_{x y }\left(1-\hat{Y}_{x y }\right)^\alpha \log \left(\hat{Y}_{x y }\right),
\vspace{-2mm}\end{equation}
\vspace{-1.4mm}
otherwise:
\vspace{1mm}
\begin{equation}
L_{shape}=-\frac{1}{N} \sum_{x y}
\left(1-Y_{x y }\right)^\beta\left(\hat{Y}_{x y }\right)^\alpha
\log \left(1-\hat{Y}_{x y }\right),
\vspace{-1.5mm}\end{equation}
where $\alpha=2$ and $\beta=4$ are default hyper-parameters and $N$ is the number of $p$ where $P(O_{s} )=1$. Since PSC module involves only fast 2D convolutions and MLPs, this guarantees detection efficiency.\vspace{-1mm}
\subsection{Attention-based Densification Fusion Module}
Some parts of object shapes significantly determine the performance and deserve more attention. Inspired by \cite{woo2018cbam} we exploit an effective hybrid-attention-based ADF module for adaptive feature refinement. The ADF module extracts shape knowledge from $\hat{S} _{BEV}$ to enhance the point cloud features ${F} _{b}$ from the backbone network. As shown in Fig.\ref{P3}, targeting the sparsity of both $\hat{S} _{BEV}$ and ${F} _{b}$, the concatenation feature ${F} _{c}$ first densifies by 2D convolutional layers making the initial dense fusion feature ${F} _{id}$ available. The ADF module consists of channel-wise attention and grid-wise attention.

Channel-wise attention focuses on filtering significant semantics in ${F} _{id}$. We first aggregate spatial information in two descriptors by using  average-pooling and max-pooling, then both descriptors are forwarded to a shared network producing a channel map. Afterwards, we merge the feature maps by summation, followed by a sigmoid function $\sigma$. In short, the computation of channel-wise attention map is: 
\vspace{-1mm}
\begin{equation}
\begin{split}
{M}_{\mathbf{c}}({F} _{id})=\sigma(MLP(F_{id}^{Avg})+MLP(F_{id}^{Max})).
\end{split}
\end{equation}

Grid-wise attention focuses on filtering significant positions in ${F} _{id}$. We apply average-pooling and max-pooling to compress the channel dimension of ${F} _{id}$, connected with a convolution layer to encode the part of shape requiring more attention. Finally, we generate a grid attention map by utilizing the inter-spatial relationship of ${F} _{id}$. The computation of grid-wise attention map is: 
\vspace{-0.5mm}
\begin{equation}
{M}_{\mathbf{g}}({F} _{id})=\sigma(f^{7 \times 7}([F_{id}^{Avg'};F_{id}^{Max'}])),
\end{equation}
where $\sigma$ denotes the sigmoid function and $f^{7 \times 7}$ represents a $7\times7$ convolutional layer.

These two attention modules focus on semantics and positions respectively. Using $\otimes$ to denote element-wise multiplication, the overall attention process can be computed as:
\vspace{-2mm}
\begin{equation}
{F}_{adf}={M}_{\mathbf{g}}({F} _{id}) \otimes {M}_{\mathbf{c}}({F} _{id}) \otimes {F} _{id}.
\end{equation}
\vspace{-7mm}

\subsection{One-Stage and Two-Stage BSH-Det3D}
Choosing to focus on accuracy or speed, the backbone network adopts two ways to encode raw points. For speed, we encode points into pillar feature maps\cite{lang2019pointpillars}. For accuracy, we encode points into voxels\cite{zhou2018voxelnet}, and then the 3D backbone extracts features by sparse convolution and compresses along the height axis to generate 2D feature maps. After encoding, the 2D backbone uses a multi-scale feature fusion network to generate a 2D feature map.

\textbf{One-Stage.} The Region Proposal Network (RPN) takes the output features of the ADF module and predicts stage-one proposals with the anchor-based approaches \cite{yan2018second, lang2019pointpillars}. Two anchors of $0^{\circ}$, $90^{\circ}$ are evaluated for each pixel of the BEV feature map. Each proposal contains eight parameters: center coordinates $\left(x_p, y_p, z_p\right)$, box size $\left(l_p, w_p, h_p\right)$, yaw rotation angle $\theta_p$, and classification confidence $c_p$.

\textbf{Two-Stage.} Based on proposals and grid features learned from stage-one, the refinement module further exploits the BEV shape heatmaps, as shown in Fig.\ref{P3}. Each box needs to consider the nearby geometry structure to generate accurate final proposals. We design a RoI-grid pooling strategy inspired by\cite{deng2021voxel}. For each proposal, we construct local grids of $6\times 6\times 6$ to capture contextual information among the neighboring voxel features. To further enhance the awareness of shape deterioration, we pool the BEV shape heatmap $\hat{S} _{BEV}$ onto the nearby grids through bilinear-interpolation and aggregate them by fusing multiple levels of ${F} _{adf}$. After that, two branches of MLP are used to predict an IoU-related class confidence score and residuals between the 3D proposal and ground truth bounding box for two-stage results.

\subsection{Loss Function}
The proposed BSH-Det3D is trained in an end-to-end manner. Our overall loss includes ${L}_{r p n}$ in stage-one, box refinement loss ${L}_{rcnn}$ in stage-two, and BEV shape heatmap estimation loss ${L}_{shape}$ in Eq.\ref{eq2} as:
\vspace{-0.5mm}
\begin{equation}\label{e7}
{L}_{\text {total }}=\lambda  {L}_{\text {shape }}+{L}_{r p n}+{L}_{p r}.
\end{equation}

Following \cite{lang2019pointpillars,yan2018second}, ${L}_{r p n}$ is defined as the summation of classification loss and box regression loss as:
\vspace{-1mm}
\begin{equation}
\begin{split}
L_{{rpn}}=&L_{{cls}}+\sum_{r}{L}_{{smooth}-\mathrm{L} 1}(\Delta\mathrm{\mathit{r_{1}}}),\\
{\mathit{r_{1}} } &\in\{x, y, z, l, h, w, \theta\}
\end{split}
\end{equation}
where $\mathit{r_{1}}$ is the parametric representation of the proposal, and the smooth-L1 loss is used to anchor box regression with the predicted residual and the regression target. We use focal loss\cite{lin2017focal} to calculate the anchor classification:
\vspace{-1mm}
\begin{equation}
{L}_{\text {cls }}=\alpha\left(1-p_t\right)^\gamma \log \left(p_t\right),
\end{equation}
where $p_t$ is the class probability of an anchor and we use the default hyper-parameters $\alpha = 0.25$ and $\gamma = 2$.

The proposal refinement loss ${L}_{rcnn}$ includes the IoU-guided confidence prediction loss\cite{shi2020pv} and box refinement loss as:
\vspace{-1mm}
\begin{equation}\vspace{-1mm}
\begin{split}
L_{{rcnn}}=&L_{{iou}}+\sum_{\mathit{r}_{2}} {L}_{{smooth}-{L} 1}\left(\Delta \mathit{r}_{2} \right),
\end{split}
\end{equation}
where $\Delta \mathit{r}_{2}$ is the residual between the predicted box and proposal target which are encoded similarly to $\Delta\mathrm{\mathit{r_{1}}}$.

\section{Experiments}
\subsection{Dataset and Evaluation}
We train and evaluate our proposed BSH-Det3D on the widely acknowledged KITTI dataset\cite{geiger2012we}, which offers 7,481 samples for training and 7,518 samples for testing. Referring to previous work such as \cite{yan2018second,shi2020pv}, we split training examples into the train set (3,712 samples) and the \textit{val} set (3,769 samples). The samples are classified as three difficulty levels: easy, moderate, and hard, the official KITTI leaderboard is ranked on the moderate levels. We adopt AP with a 3D overlap threshold of 0.7 as the evaluation metric of the Car class and 0.5 for Cyclist.

\begin{table*}[]
\tabcolsep=0.28cm
\begin{center}
\caption{Performance comparison of object detection with SOTA LiDAR methods of KITTI \textit{val} split.}\label{val}
\renewcommand\arraystretch{1.2}
\begin{tabular}{c|c|ccc|ccc|ccc|cll}
\hline\hline
                            &                                     & \multicolumn{3}{c|}{Car 3D $AP_{R40}$}                                                                                    & \multicolumn{3}{c|}{Car BEV $AP_{R40}$}                                                                                   & \multicolumn{3}{c|}{Cyc. 3D $AP_{R40}$}                                                                                   & \multicolumn{3}{c}{}                                                                      \\
\multirow{-2}{*}{Stage}     & \multirow{-2}{*}{Method}            & Easy                                   & \underline{ Mod.}                                   & Hard                                   & Easy                                   & \underline{ Mod.}                                   & Hard                                   & Easy                                   & \underline{ Mod.}                                   & Hard                                   & \multicolumn{3}{c}{\multirow{-2}{*}{\begin{tabular}[c]{@{}c@{}}Time\\ (ms)\end{tabular}}} \\ \hline
                            & PointPillars\cite{lang2019pointpillars}                        & 87.75                                  & 78.39                                  & 75.18                                  & 92.40                                  & 87.79                                  & 86.39                                  & 81.57                                  & 62.94                                  & 58.98                                  & \multicolumn{3}{c}{24}                                                                    \\
                            & BSH-Det3D(Pillars)                    & 89.08                                  & 81.66                                  & 79.01                                  & 92.80                                  & 89.15                                  & 88.46                                  & 86.48                                      & 69.22                                      & 63.59                                      & \multicolumn{3}{c}{32}                                                                    \\
                            & \cellcolor[HTML]{E4F9F9}Improvement & \cellcolor[HTML]{E4F9F9}\textbf{+1.33} & \cellcolor[HTML]{E4F9F9}\textbf{+3.27} & \cellcolor[HTML]{E4F9F9}\textbf{+3.83} & \cellcolor[HTML]{E4F9F9}\textbf{+0.40} & \cellcolor[HTML]{E4F9F9}\textbf{+1.36} & \cellcolor[HTML]{E4F9F9}\textbf{+2.07} & \cellcolor[HTML]{E4F9F9}\textbf{+4.91} & \cellcolor[HTML]{E4F9F9}\textbf{+6.28} & \cellcolor[HTML]{E4F9F9}\textbf{+4.61} & \multicolumn{3}{c}{\cellcolor[HTML]{E4F9F9}-8}                                            \\ \cline{2-14} 
                            & SECOND\cite{yan2018second}                              & 90.97                                  & 79.94                                  & 77.09                                  & 95.61                                  & 89.54                                  & 86.96                                  & 78.50                                  & 56.74                                  & 52.83                                  & \multicolumn{3}{c}{50}                                                                    \\
                            & BSH-Det3D(Voxels)                     & 91.07                                  & 82.53                                  & 79.54                                  & 93.04                                  & 89.28                                  & 88.43                                  & 85.32                                  & 66.23                                  & 64.92                                  & \multicolumn{3}{c}{43}                                                                    \\
\multirow{-6}{*}{One-stage} & \cellcolor[HTML]{E4F9F9}Improvement & \cellcolor[HTML]{E4F9F9}\textbf{+0.10} & \cellcolor[HTML]{E4F9F9}\textbf{+2.59} & \cellcolor[HTML]{E4F9F9}\textbf{+2.45} & \cellcolor[HTML]{E4F9F9}-2.57          & \cellcolor[HTML]{E4F9F9}-0.26          & \cellcolor[HTML]{E4F9F9}\textbf{+3.27} & \cellcolor[HTML]{E4F9F9}\textbf{+6.82} & \cellcolor[HTML]{E4F9F9}\textbf{+9.49} & \cellcolor[HTML]{E4F9F9}\textbf{+12.09} & \multicolumn{3}{c}{\cellcolor[HTML]{E4F9F9}\textbf{+7}}                                   \\ \hline
                            & PV-RCNN\cite{shi2020pv}                             & 92.10                                  & 84.36                                  & 82.48                                  & 93.02                                  & 90.33                                  & 88.53                                  & 88.88                                  & 71.95                                  & 66.78                                  & \multicolumn{3}{c}{80}                                                                    \\
                            & Voxel R-CNN\cite{deng2021voxel}                         & 92.38                                  & 85.29                                  & 82.86                                  & 95.52                                  & 91.25                                  & 88.99                                  & -                                      & -                                      & -                                      & \multicolumn{3}{c}{40}                                                                    \\
                            & BSH-Det3D(Refinement)                 & 92.88                                  & 85.93                                  & 83.49                                  & 93.97                                  & 91.94                                  & 89.60                                  & 89.64                                      & 72.33                                      & 69.05                                      & \multicolumn{3}{c}{48}                                                                    \\
\multirow{-4}{*}{Two-stage} & \cellcolor[HTML]{E4F9F9}Improvement & \cellcolor[HTML]{E4F9F9}\textbf{+0.50} & \cellcolor[HTML]{E4F9F9}\textbf{+0.64} & \cellcolor[HTML]{E4F9F9}\textbf{+0.63} & \cellcolor[HTML]{E4F9F9}-1.55          & \cellcolor[HTML]{E4F9F9}\textbf{+0.69} & \cellcolor[HTML]{E4F9F9}\textbf{+0.61} & \cellcolor[HTML]{E4F9F9}\textbf{+0.76} & \cellcolor[HTML]{E4F9F9}\textbf{+0.38}              & \cellcolor[HTML]{E4F9F9}\textbf{+2.27}              & \multicolumn{3}{c}{\cellcolor[HTML]{E4F9F9}-8}                                            \\ \hline
\end{tabular}
\end{center}
\vspace{-6mm}
\end{table*}

\subsection{Implementation Details}
\textbf{Pillarization\&Voxelization.} Before sending to networks, the raw points are first encoded into pillars or voxels. For voxelization, we clip the range of point clouds into $[0,70.4]m$ for the X-axis, $[-40,40]m$ for the Y-axis, and $[-3,1]m$ for the Z-axis. The input voxel size is set as $(0.05m, 0.05m, 0.1m)$. For pillarization, we define the detection range as $[0,69.12]m$ for the X-axis, $[-39.68,39.68]m$ for the Y-axis, and $[-3,1]m$ for the Z-axis. We set the pillar size to $(0.16m,0.16m,4m)$.

\textbf{Training.} The BSH-Det3D is end-to-end optimized by the ADAM optimizer\cite{kingma2014adam} from scratch. The parameter $\lambda$  in Eq.\ref{e7} is set to 6.0 empirically. We train our models with a batch size of 8 on a GTX 3090 GPU for 80 epochs. The learning rate is initialized as 0.01 and updated by the cosine annealing strategy. We randomly sample 128 proposals for training, and $50\%$ of them are positive samples that have IoU$>$0.55 with the corresponding ground truth boxes. 

\textbf{Inference.} During the inference stage, non-maximum suppression (NMS) is conducted with a threshold$>$0.7 to filter the redundant proposals. We choose the top 100 proposals for refinement.  After refinement, NMS is applied with IoU threshold 0.01 to remove redundant box predictions.

\textbf{Data augmentation.} First, we randomly sample objects from the training data and inject them into the training samples as \cite{yan2018second}. Next, we randomly flip scenes along X-axis with $50\%$ probability. Then, we rotate each scene around Z-axis with a random angle sampled from $[-45^{\circ},45^{\circ}]$. Finally, we uniformly sample a scaling factor from the range of $[0.95, 1.05]$ and use it to scale the point cloud.

\begin{table*}[]
\tabcolsep=0.4cm
\begin{center}
\caption{Performance comparison of 3D and BEV detection of car class on KITTI test split.}\label{test}
\renewcommand\arraystretch{1.2}
\begin{tabular}{c|c|c|ccc|ccc|c}
\hline\hline
\multirow{2}{*}{Modality}    & \multirow{2}{*}{Method} & \multirow{2}{*}{Stage} & \multicolumn{3}{c|}{3D Detection $AP_{R40}$} & \multicolumn{3}{c|}{BEV Detection $AP_{R40}$} & \multirow{2}{*}{\begin{tabular}[c]{@{}c@{}}Time\\ (ms)\end{tabular}} \\
                             &                         &                        & Easy      & \underline{ Mod.}     & Hard     & Easy      & \underline{ Mod.}     & Hard      &                                                                      \\ \hline
\multirow{7}{*}{RGB+LiDAR}   & MV3D\cite{chen2017multi}                    & Two                    & 74.97     & 63.63          & 54.00    & 86.62     & 78.93          & 69.80     & 360                                                                  \\
                             & AVOD\cite{ku2018joint}                    & Two                    & 83.07     & 71.76          & 65.73    & 89.75     & 84.95          & 78.32     & 100                                                                        \\
                             & ContFuse\cite{liang2018deep}                & One                    & 83.68     & 68.78          & 61.67    & 94.07     & 85.35          & 75.88     & 60                                                                   \\
                             & UberATG-MMF\cite{liang2019multi}             & Two                    & 88.40     & 77.43          & 70.22    & 93.67     & 88.21          & 81.99     & 80                                                                   \\
                             & 3D-CVF\cite{yoo20203d}                  & Two                    & 89.20     & 80.05          & 73.11    & 93.52     & 89.56          & 82.45     & 75                                                                   \\
                             & CLOCs PVCas\cite{pang2020clocs}             & Two                    & 88.94     & 80.67          & 77.15    & 93.05     & 89.80          & 86.57     & 100                                                                  \\ \hline
\multirow{15}{*}{LiDAR only} & SECOND\cite{yan2018second}                  & One                    & 83.34     & 72.55          & 65.82    & 89.39     & 83.77          & 78.59     & 50                                                                   \\
                             & PointPillars\cite{lang2019pointpillars}            & One                    & 82.58     & 74.34          & 68.99    & 90.07     & 86.56          & 82.81     & \textbf{24}                                                                   \\
                             & PointRCNN\cite{shi2019pointrcnn}               & Two                    & 86.96     & 75.64          & 70.70    & 92.13     & 87.39          & 82.72     & 100                                                                               \\
                             & STD\cite{yang2019std}                     & Two                    & 87.95     & 79.71          & 75.09    & 94.74     & 89.19          & 86.42     & 80                                                                   \\
                             & Part-A2\cite{shi2020points}                 & Two                    & 87.81     & 78.49          & 73.51    & 91.70     & 87.79          & 84.61     & 80                                                                   \\
                             & Associate-3Det\cite{du2020associate}          & One                    & 85.99     & 77.40          & 70.53    & 91.40     & 88.09          & 82.96     & 60                                                                   \\
                             & 3DSSD\cite{yang20203dssd}                   & One                    & 88.36     & 79.57          & 74.55    & 92.66     & 89.02          & 85.86     & 38                                                                   \\
                             & PV-RCNN\cite{shi2020pv}                 & Two                    & 90.25     & 81.43          & 76.82    & \textbf{94.98}     & 90.65          & 86.14     & 80                                                                                                             \\
                             & Point-GNN\cite{shi2020point}               & One                    & 88.33     & 79.47          & 72.29    & 93.11     & 89.17          & 83.90     & 643                                                                  \\
                             & TANet\cite{liu2020tanet}                   & One                    & 84.39     & 75.94          & 68.82    & 91.58     & 86.54          & 81.19     & 35                                                                   \\
                             & Voxel R-CNN\cite{deng2021voxel}             & Two                    & \textbf{90.90}     & 81.62          & 77.06    & 94.85     & 88.83          & 86.13     & 40                                                                   \\
                             & CIA-SSD\cite{zheng2021cia}                 & One                    & 89.59     & 80.28          & 72.87    & 93.74     & 89.84          & 82.39     & 31                                                                   \\ \cline{2-10} 
                             & BSH-Det3D(ours)         & Two                    & 88.75     & \textbf{81.91}          & \textbf{77.36}    & 92.90     & \textbf{90.99}          & \textbf{86.43}     & 48\\ \hline
\end{tabular}
\end{center}
\vspace{-4mm}
\end{table*}
\subsection{Evaluation Results}
We evaluate BSH-Det3D for 3D detection and BEV detection benchmark on KITTI \textit{val} split and test split. Corresponding to the KITTI\cite{geiger2012we} protocol, we calculate the AP results under 40 recall thresholds (R40). The evaluation results show three major advantages of our method.

\textbf{High performance.} 
We compare our two-stage detector BSH-Det3D(Refinement) with the front runners on the KITTI leaderboard by submitting our results to the online test server, as illustrated in Table \ref{test}. Our BSH-Det3D can effectively improve detection performance and achieves balance between accuracy and efficiency. By taking full advantage of BEV shape knowledge, BSH-Det3D(Refinement) achieves 81.91\% average precision on the moderate level of class Car with 48ms. Our method outperforms many multi-modality fusion-based methods, including UberATG-MMF\cite{liang2019multi}, 3D-CVF\cite{yoo20203d}, and CLOCs PVCas\cite{pang2020clocs}, by a large margin (1.24\% to 4.48\% of moderate AP and 0.21\% to 7.14\% of hard AP). Compared with the LiDAR-based methods, we also outperform the recent SOTA detectors, \textit{e.g.}, PV-RCNN\cite{shi2020pv}, Voxel R-CNN\cite{deng2021voxel}, and CIA-SSD\cite{zheng2021cia} by 0.29\% to 1.63\% of moderate AP and 0.30\% to 4.49\% of hard AP. The KITTI test set results demonstrate that our proposed BSH-Det3D achieves the SOTA performance on 3D object detection and keeps the high efficiency to address shape deterioration.

\textbf{Flexibility.} As summarized in Table \ref{val}, to demonstrate that BSH-Det3D can generalize across models, we build BSH-Det3D using two mainstream detectors\cite{lang2019pointpillars}, \cite{yan2018second} implemented in OpenPCDet\cite{openpcdet2020}. The box refinement module is designed based on the SECOND detector \cite{yan2018second}. We evaluate BSH-Det3D on Car and Cyclist categories and use the $AP_{R40}$ for PV-RCNN\cite{shi2020pv} and Voxel R-CNN\cite{deng2021voxel} from their papers and the $AP_{R40}$ for PointPillars\cite{lang2019pointpillars} and SECOND\cite{yan2018second} are generated from the officially released code. Compared to baseline methods, the  BSH-Det3D detectors increase the performance in 3D and BEV object detection by a large margin. Results demonstrate the effectiveness and flexibility of our method. 

\begin{figure}[ht]
	\centering
	\includegraphics[scale=0.43]{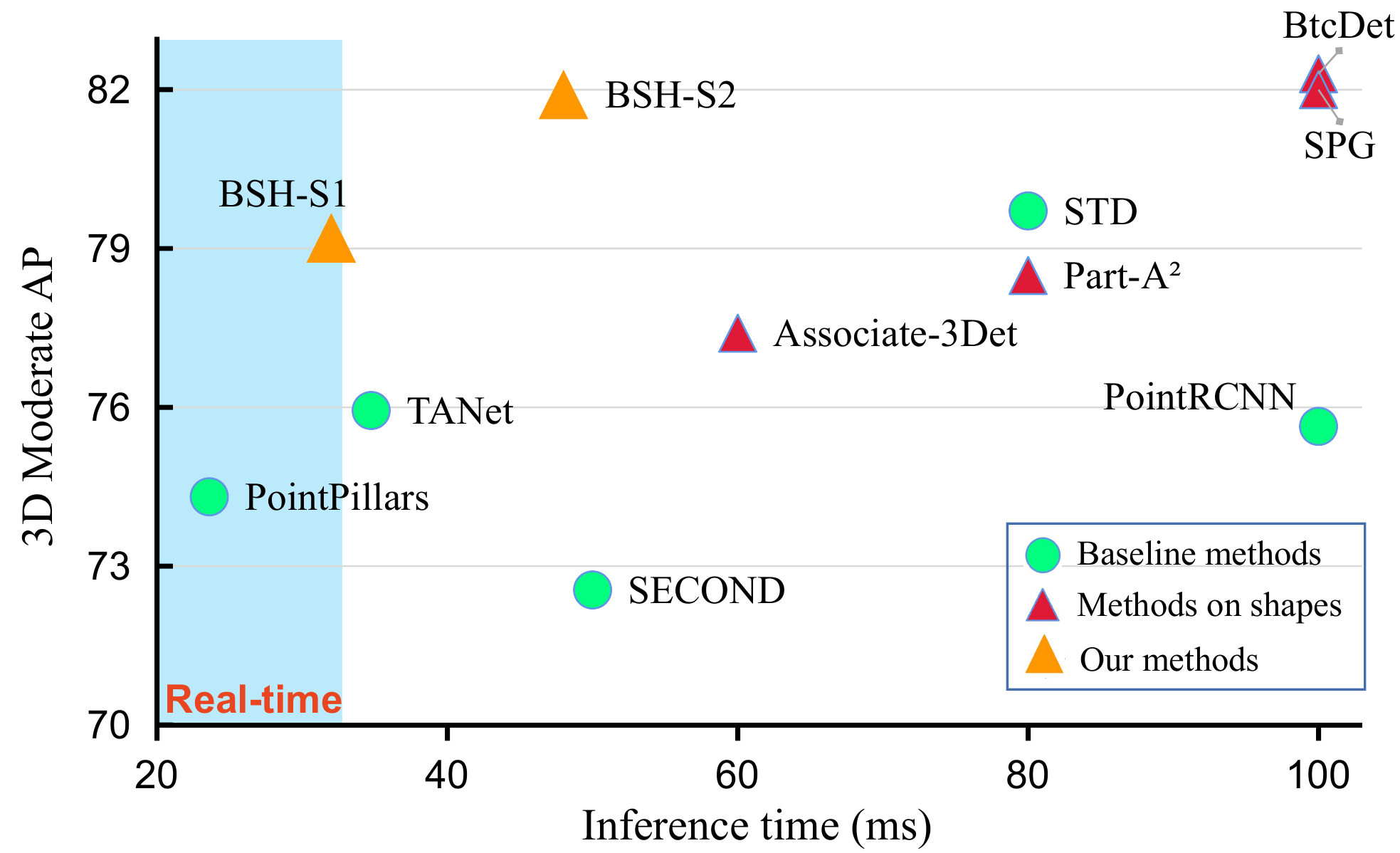}
	\vspace{-7.4mm}
	\caption{The 3D detection performance and speed for our BSH-Det3D on the KITTI test set compared with SOTA detectors, especially those focusing on shapes missing. BSH-S1 and BSH-S2 show the result of BSH-Det3D(Pillars) and BSH-Det3D(Refinement), respectively.}\label{P8}\vspace{-6mm}
\end{figure}

\textbf{Efficiency.} Notably, owing to the fast 2D CNN architecture in PSC module and the effective fusion strategy in ADF module, our method is quite efficient (Table \ref{val}). We measure the runtime of BSH-Det3D on an Intel i7-12700F and a single 3060Ti GPU. The moderate AP and running speed comparisons in the KITTI server are shown in Fig.\ref{P8}. 

Comparing to other detectors focus on shape missing, our one-stage detector BSH-Det3D(Pillars) is the only detector that works in real-time (32ms). It also achieves 79.10\% AP, significantly improving the baseline PointPillars \cite{lang2019pointpillars} by 4.76\% of moderate AP. Moreover, BSH-Det3D(Refinement) achieves comparable accuracy with the strong competitors, \textit{i.e.}, BtcDet\cite{xu2022behind} and SPG\cite{xu2021spg}, with only about 50\% of the running time. This verifies that the BEV shapes are almost sufficient for shape deterioration in 3D object detection, and the BEV shape representation is more efficient than specific 3D shapes.

\begin{figure*}[ht]
	\centering
	\includegraphics[scale=0.9]{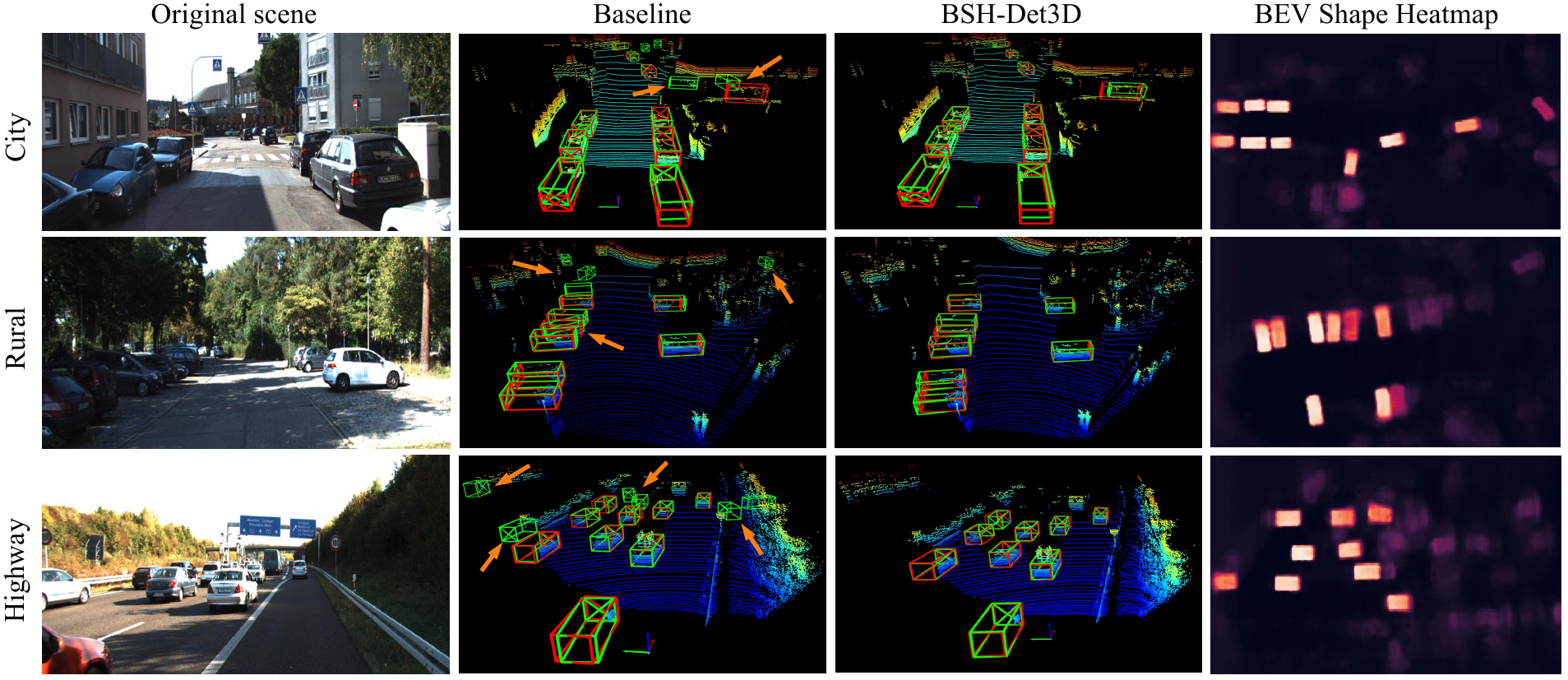}
	\vspace{-3mm}
	\caption{Qualitative results of the KITTI \textit{val} split. We select three typical scenes in KITTI: City, Rural, and Highway. We show the BSH-Det3D(Voxels) results with corresponding BEV shape heatmaps for each scene. The advantages compared with the baseline\cite{yan2018second} are marked with the orange arrows. The boxes of prediction and ground truth are rendered in green and red. Results show that our method positively reduces false detections and corrects offset.
}
	\label{P6}
	\vspace{-2mm}
\end{figure*}

\textbf{Qualitative results.} Some visualization results for BSH-Det3D(Voxels) and the corresponding BEV shape heatmaps are illustrated in Fig.\ref{P6}. We show detection results on three typical KITTI scenes: City, Rural, and Highway. In comparison to the baseline\cite{yan2018second} method, we mark the advantage of BSH-Det3D with orange arrows. In the city, objects shapes are obscured at corner intersections, which causes missed detection in the baseline. However, due to the guidance of the BEV shape, the missing shapes are completed with shape heatmaps, and all objects can get effective detection in BSH-Det3D. In rural areas, the vehicles on the left are parked compactly, and the baseline\cite{yan2018second} suffers detection drift by limitation of points information. In comparison, by taking pillar-wise shape estimation, shapes can be separated under compact parking areas, and the drift box can be corrected. Regarding highways, the vehicles driving side by side can cause multiple occlusions, resulting in the baseline method facing critical false detection, especially in far-range areas. By associating shape heatmaps, our detector can provide more descriptions of objects, significantly alleviating false detections. In conclusion, BSH-Det3D has benefits in various scenes, effectively suppressing missing and false detections.
\subsection{Ablation Studies}
\textbf{Effect of Components.} Table \ref{abl} details how each proposed module influences the accuracy and efficiency of our BSH-Det3D. The results are evaluated with $AP_{R40}$ of moderate level for the car class. \textit{Method(a)} is the one-stage baseline that performs detection on BEV features which runs at 25.5ms. \textit{Method(b)} extends \textit{(a)} with a pillar-based shape completion module, which directly concatenates with raw points feature and fusion by 2D convolution. The PSC module leads to a boost of 1.79\% moderate AP, which verifies that PSC can strengthen the robustness of spatial features. We apply pillarization and 2D CNN leading to a decrease of 33.2ms. \textit{Method(c)} replaces feature fusion with our ADF module, which makes 1.11\% AP improvement and only an extra 2.4ms thanks to a simple yet effective Hybird-Attention feature fusion strategy. \textit{Method(d)} uses our rendered Gaussian kernels (GAU) to counteract the noise of shape priors during training, boosting 0.33\% AP without extra cost. \textit{Method(e)} is the proposed BSH-Det3D(Refinement) by extending the one-stage detector with a box refinement module combined with BEV shape knowledge. BSH-Det3D(Refinement) achieves SOTA accuracy for 3D object detection and maintains efficiency.
 \begin{table}[]
 \tabcolsep=0.2cm
\caption{Performance with different configurations on KITTI \textit{val} set.}\label{abl}\vspace{-1mm}
\renewcommand\arraystretch{1.1}
\begin{tabular}{c|cccc|c|c}
\hline\hline
Methods & PSC & GAU & ADF & REF & 3D $AP_{R40}$ & Time (ms) \\ \hline
(a)     &     &     &     &     & 79.94         & 25.5    \\
(b)     & \checkmark   &     &     &     & 81.73         & 33.2    \\
(c)     & \checkmark   &     & \checkmark   &     & 82.84         & 35.6    \\
(d)     & \checkmark   & \checkmark   & \checkmark   &     & 83.17         & 35.6    \\
(e)     & \checkmark   & \checkmark   & \checkmark   & \checkmark   & 85.93         & 48.2    \\ \hline
\end{tabular}
    \vspace{-6mm}
\end{table}

\begin{table}[]
\tabcolsep=0.25cm
\caption{Detection of different intervals on KITTI \textit{val} set.}\label{interval}
    \vspace{-1mm}
\renewcommand\arraystretch{1.5}
\begin{tabular}{c|c|ccl|c}
\hline\hline
\multirow{2}{*}{\begin{tabular}[c]{@{}c@{}}Recall\\ Interval\end{tabular}} & \multirow{2}{*}{Method} & \multicolumn{3}{c|}{Car 3D Detection} & \multirow{2}{*}{Ration (\%)} \\ \cline{3-5}
                         &              & \multicolumn{1}{c|}{TP}                    & \multicolumn{2}{c|}{FP}   &       \\ \hline
\multirow{2}{*}{R1-R10}  & PointPillars\cite{lang2019pointpillars} & \multicolumn{1}{c|}{\multirow{2}{*}{1772}} & \multicolumn{2}{c|}{23}   & 98.72 \\ \cline{2-2} \cline{4-6} 
                         & BSH-Det3D      & \multicolumn{1}{c|}{}                      & \multicolumn{2}{c|}{17}   & 99.04 (0.32$\uparrow$) \\ \hline
\multirow{2}{*}{R11-R20} & PointPillars\cite{lang2019pointpillars} & \multicolumn{1}{c|}{\multirow{2}{*}{3740}} & \multicolumn{2}{c|}{157}  & 95.97 \\ \cline{2-2} \cline{4-6} 
                         & BSH-Det3D      & \multicolumn{1}{c|}{}                      & \multicolumn{2}{c|}{124}  & 96.80 (0.83$\uparrow$) \\ \hline
\multirow{2}{*}{R21-R30} & PointPillars\cite{lang2019pointpillars} & \multicolumn{1}{c|}{\multirow{2}{*}{5709}} & \multicolumn{2}{c|}{767}  & 88.16 \\ \cline{2-2} \cline{4-6} 
                         & BSH-Det3D      & \multicolumn{1}{c|}{}                      & \multicolumn{2}{c|}{600}  & 90.49 (2.33$\uparrow$)\\ \hline
\multirow{2}{*}{R31-R40} & PointPillars\cite{lang2019pointpillars} & \multicolumn{1}{c|}{\multirow{2}{*}{6496}} & \multicolumn{2}{c|}{3567} & 64.55 \\ \cline{2-2} \cline{4-6} 
                         & BSH-Det3D      & \multicolumn{1}{c|}{}                      & \multicolumn{2}{c|}{1836} & 77.96 (\textbf{13.41$\uparrow$}) \\ \hline
\end{tabular}
    \vspace{-5.0mm}
\end{table}
 
\textbf{Effect on different recall thresholds.} We designed an experiment based on four intervals to further analyze how BSH-Det3D enhances the detectors. Since the recall positions of R40 are recorded according to the sorted predicted confidence, objects in $[R1, R30]$ generally have low detection difficulty and retain almost complete object shapes. In comparison, the objects of $[R31, R40]$ usually suffer from shape deterioration in occluded and distant areas.
Thus, our experiment evenly slices the 40 recall points into four pieces, counts the number of true positives (TP) and false positives (FP), then calculates the TP's ratios (TP/(TP+FP)) in each interval. Our experiment is conducted on BSH-Det3D(Voxels) of the KITTI \textit{val} set, moderate level on Car category. As demonstrated in Table \ref{interval}, for $[R1, R30]$, the TP ratio of our methods has a slight improvement over the baseline. However, for $[R31, R40]$, the ratio of FP is greatly reduced in BSH-Det3D, which plays a significant role in detection performance. It can be inferred that with the associated BEV shape heatmap, BSH-Det3D can provide additional descriptions of objects, which is especially useful in correcting low-confidence boxes that suffer from shape deterioration.
\section{Conclusion}    
In this paper, we demonstrate that shape deterioration is a fundamental challenge in 3D object detection. To tackle this, we present the novel BSH-Det3D exploiting the potential of BEV shapes to improve detection. Specifically, we design a efficient PSC module that learns to enhance spatial features by producing the complete BEV shapes in each scene. Additionally, we introduce a hybrid-attention based ADF module for adaptive feature refinement between shapes and raw points with negligible overhead. It should also be noted that the BEV shape heatmaps of our approach can be easily integrated into many existing detectors in 3D point clouds. Experimental results on the KITTI benchmark dataset have validated the efficiency and flexibility of our BSH-Det3D. In future work, we plan to integrate the BEV shape completion module with RGB images to further improve the performance.
\bibliographystyle{IEEEtran}
\bibliography{ref}

%%%%%%%%%%%%%%%%%%%%%%%%%%%%%%%%%%%%%%%%%%%%%%%%%%%%%%%%%%%%%%%%%%%%%%%%%%%%%%%%

%%%%%%%%%%%%%%%%%%%%%%%%%%%%%%%%%%%%%%%%%%%%%%%%%%%%%%%%%%%%%%%%%%%%%%%%%%%%%%%%

\end{document}